\documentclass{article}

\usepackage{arxiv}

\usepackage[utf8]{inputenc} 
\usepackage[T1]{fontenc}    
\usepackage{hyperref}       
\usepackage{url}            
\usepackage{booktabs}       
\usepackage{amsfonts}       
\usepackage{nicefrac}       
\usepackage{microtype}      
\usepackage{lipsum}		
\usepackage{graphicx}
\usepackage{doi}

\usepackage[numbers]{natbib}

\title{On learning spatial sequences with the movement of attention}

\author{\href{https://orcid.org/0000-0002-6933-1217}{\includegraphics[scale=0.06]{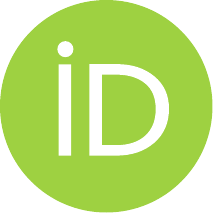}\hspace{1mm}V.M. Osaulenko} 
\thanks{Use footnote for providing further
		information about the author (webpage, alternative
		address)---\emph{not} for acknowledging funding agencies.} \\
	Igor Sikorsky Kyiv Polytechnic Institute \\
    Kyiv, Ukraine
\\
	\texttt{osaulenko.v.m@gmail.com} \\
}

\renewcommand{\shorttitle}{On learning spatial sequences }

\hypersetup{
pdftitle={On spatial sequence learning with the movement of attention},
pdfsubject={q-bio.NC, q-bio.QM},
pdfauthor={V.M.Osaulenko},
pdfkeywords={spatial sequence, sequence learning, movement of attention, selectionism},
}

\begin{document}
\maketitle

\begin{abstract}
In this paper we start with a simple question, how is it possible that humans can recognize different movements over skin with only a prior visual experience of them? 
 Or in general, what is the representation of spatial sequences that are invariant to scale, rotation, and translation across different modalities?
 To answer, we rethink the mathematical representation of spatial sequences, argue against the minimum description length principle, and focus on the movements of attention. 
 We advance the idea that spatial sequences must be  represented on different levels of abstraction, this adds redundancy but is necessary for recognition and generalization. 
 To address the open question of how these abstractions are formed we propose two hypotheses:
 the first invites exploring selectionism learning, instead of finding parameters in some models; 
 the second proposes to find new data structures, not neural network architectures, to efficiently store and operate over redundant features to be further selected.
 Movements of attention are central to human cognition and lessons should be applied to new better learning algorithms.

\end{abstract}


\section{Introduction}
This paper deals with the problem of learning spatial sequences like the movements of objects.
In contrast to temporal sequences, like music or text, spatial sequences have distance and direction relations for any two elements.
Interestingly, there are no senses for distances and directions but biological organisms receive this information with a sequential, not parallel, process of eye saccades and covert movements of attention. 

 Such a sequential manner drastically differs from many computational approaches, like recurrent neural networks \cite{Graves2013} that treat  the spatial sequence as temporal. Also, convolutional neural networks despite showing great performance in computer vision tasks have problems with geometry-related tasks \cite{sattler2019understanding, sarlin2021back}.
 That is why practical applications of SLAM in robotics often use hybrid approaches \cite{MacarioBarros2022}, combining neural networks with precise algorithms for spatial data.   
 
 In recent studies, it was shown that humans leverage spatial relationships for sequence learning \cite{wang2019representation, AlRoumi2021}. We discover regularities on multiple nested levels similar to the language structure. Also, we chunk long sequences into the composite to reduce the working memory load and improve the prediction of the next locations. All this is done in an unsupervised manner, with the proposed goal to reduce the minimum description length (MDL). The brain-inspired mechanisms still have a lot to share to improve the current algorithms. For example, recently it was proposed \cite{Davison2018} to return to the active vision approach for robotics perception that computes sequentially where to look for the most informative places. 
  
The goal of this paper is to highlight the main challenges in spatial sequence learning and bring new or revive old insights and ideas by analyzing human perception and mathematical representation of spatial sequences.   
The paper proceeds with a thought experiment that addresses the invariant recognition of spatial sequences. 
Next, we clarify what sequence learning is and how spatial sequences are different.
Further, we argue that most mathematical forms of sequences should be changed to better reflect invariances.
Thus, we favor the instruction form, that links with the previous research on MDL representation.
However, we provide arguments that MDL is not compatible with making abstractions, redundancy should be added to receive generalization.  
In the last section, we address the movement of attention both in space and scale. Unlike temporal sequences, spatial relationships alleviate the problem of long-range sequence dependencies and are represented on different abstraction levels. 

In the end, we propose two hypotheses, on how to proceed with brain-inspired algorithms for spatial sequence learning. 
First, invites exploring selectionism learning, instead of finding parameters in some models. The second addresses the problem of the exponential explosion of features to be selected, with the idea to explore new data structures, that are not necessarily neural network based.

\section{Thought experiment}
\begin{figure}
	\centering
    \includegraphics[width=\textwidth]{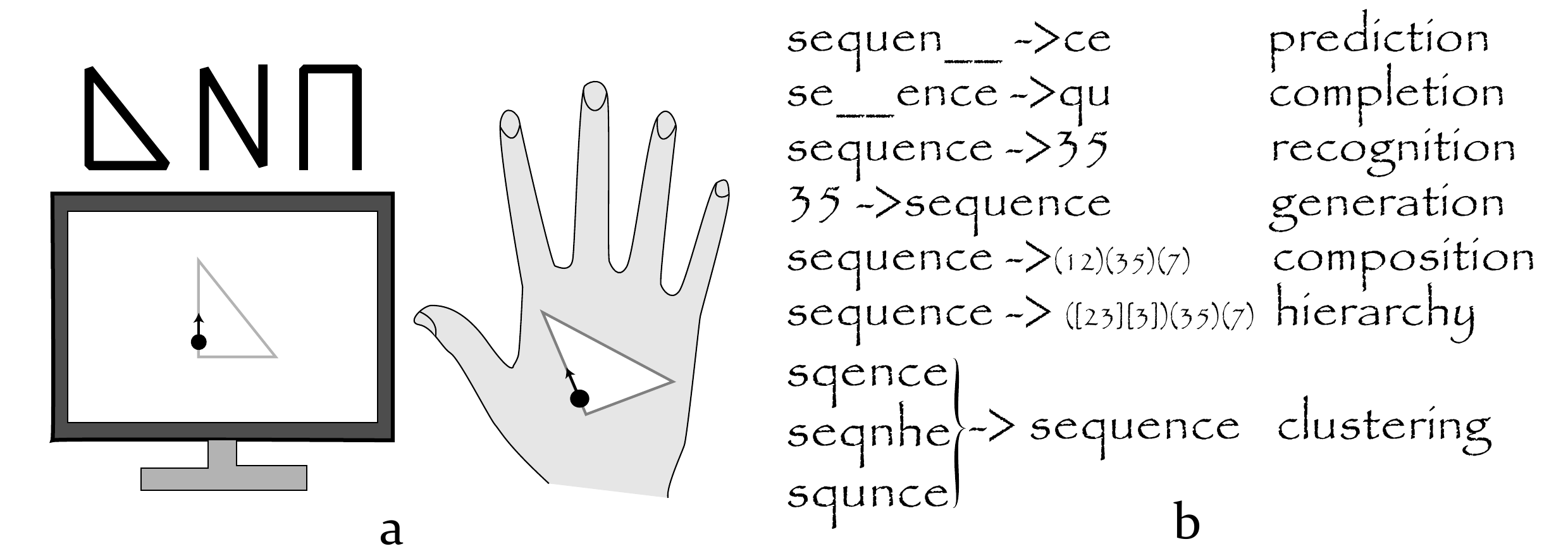}\hfill
	\caption{
 a)Illustration of a thought experiment.
 Three spatial sequences are presented visually but can be recognized with somatosensory input
 b)Summary of the main sequence learning problems
 }
	\label{fig:thought_exp}
\end{figure}

Imagine a person that looks at a screen where the black dot moves in three sequential patterns (a right triangle, letter N, and letter U).
Afterward, an experimenter draws these patterns over the skin of a person’s arm while the person’s eyes are closed (see fig. \ref{fig:thought_exp}a). 
Everyday experience tells that, trivially, the person can recognize such simple sequences over skin flawlessly. 
We can make a number of observations and conclusions from the experiment. 

1. \textit{Movement of attention}.
Even though visual information is  processed in the visual cortex and the skin perception is somatosensory, none of them are responsible for sequence recognition and prediction alone.
Otherwise, transfer of recognition between modalities would not be possible.
One could argue that both visual and somatosensory cortices are topographically organized, and serve as a saliency map,  thus both elicit similar movements of attention.  
However, drawing patterns on two arms stacked together also gives correct recognition, even though the activation is discontinuous in the somatosensory cortex. Therefore, both sequences are encoded in the same neural structure (possibly that represents the space around the body), which gives a similar continuous movement of attention in both tasks. 

 2. \textit{Invariance}. Recognition does not depend on the location where the pattern was drawn on the body. The arm, leg, or tongue does not matter as long as the movements are made sufficiently large to differentiate. The same is true for orientation and the scale of the sequences.  
 
3. \textit{Working memory}. Awareness of the location of the beginning of the sequence and its corners persists even after the stimulus presentation. 
Spatial sequences leave a lasting activity in contrast to pure temporal sequences like music or language.
Therefore, the order of the sequence has less influence on the recognition. 

4. \textit{Multiple predictions}. Before the sequence is finished, once it is recognized, it can be predicted. In case of ambiguity, multiple predictions are made. However, this relies on memory, for a more complex stimuli  prediction would be harder. 

5. \textit{Abstraction}. Recognition does not depend on noisy variations like the length or angle of the segments of patterns, although variations are noticeable. Therefore, the sequence is represented in an abstract way while being able to differentiate different instances. 

This thought experiment is an instance of what is known in the literature as  a cross-modal transfer of knowledge.
Multiple experiments showed that learning in one modality enhances or transfers to another, like for visual-auditory \cite{Yildirim2015}, visual-haptic \cite{yildirim2013transfer}, or even in motor sequences effector1-effector2 (movement of fingers and elbows in \cite{Cohen1990}). 
Based on this, the multisensory hypothesis states that an abstract amodal representation of objects must be created that is responsible for such a transfer.
Also, this idea of amodal abstractions becomes more salient in light of the recent trend for multimodal models \cite{kosmos}.

All these observations form the necessary features of any algorithm designed for spatial sequence learning. 
Note, spatial sequence processing is closely related to space representation in the entorhinal-hippocampal network, where grid cells, place cells, and many others were found, see review \cite{Whittington2022}. 
However, in this paper, we avoid biological realism and focus on a mathematical representation of spatial sequences
and the computational problems of recreating these observations.
But at first, we briefly overview the task of sequence learning. 

\section{Background in sequence learning}
A discrete sequence $s_{1:N} = \{s_i \in A, i=1:N\}$ of length $N$ is an ordered collection of elements from some set $A$. 
Individual elements are not comparable, but the sequence induces the temporal structure, like $s_i$ and $s_{i+1}$ are adjacent in time. 
Usually, the temporal structure is described by a joint probability distribution over all elements $p(s_{1:N})$. 
The next categorization is based on \cite{sun2001sequence}, with some modification and extension.  

Sequence learning includes many related tasks over sequences. 
The central is a \textit{sequence prediction}, where given a sequence $s_{1:n}$ the task is to predict the next $k$ elements $s_{n+1:n+k}$.
This is closely related to \textit{pattern completion}, with the difference that the task includes outputting missing elements in the middle or at the beginning of a sequence.  
\textit{Sequence recognition} deals with the problem of assigning the correct label to a sequence $s_{1:n} -> l$, where the label can be just a number. 
In \textit{online sequence recognition} the output is a distribution of possible labels after the presentation of elements one by one.
\textit{Sequence generation} is a reverse problem of retrieving the sequence from a label ($l->s_{1:m}$). 
We can use online recognition and generation to solve the prediction task. 
If recognition outputs multiple hypotheses, then multiple predictions are made. 
\textit{Sequence-to-sequence} learning links one sequence to another $s^1_{1:n}->s^2_{1:m}$, which is common in practical applications, like text translation.
Great importance has a \textit{sequence clustering} task with the goal to find groups of similar sequences, especially for sequences with complex structure, noise, and variations. 
Clustering reduces dimensions and is also known as finding the prototype or forming the abstraction. 
Closely related to clustering is a \textit{sequence compression} task. Lossless compression reduces the redundancy of temporal structure but does not lose information and is widely used for text or code compression. Lossy sequence compression discards less useful information based on some criteria, thus compresses more but cannot restore the original sequence. 

To reduce dimensionality it is essential to utilize the structure of the sequence.
The sequence has a \textit{compositional structure} if it is formed from other sequences  $s_{1:n}=s^1_{1:k}s^2_{1:m}$.
Note, any sequence can be split into subsequences arbitrarily, however, a good division utilize the statistics where the resulted subsequences appear in many other sequences. 
The sequence has a \textit{hierarchical structure} if it has a recursive compositional structure, where the subsequences are composed of other subsequences. See fig.\ref{fig:thought_exp}b which illustrates major sequence learning tasks together.
One of the prominent ideas is that by finding the minimum description length (MDL) of the sequence (that is related to Kolmogorov complexity) the algorithm should uncover the hierarchical and compositional structure. However, no such algorithm exists to this day, also with some notable attempts \cite{franz2019william}.

This paper considers spatial sequences where each element belongs to k-dimensional space $r_{1:N}= \{(r_i): r_i \in R^k, i=1:N\}$. 
Usually, the space is Euclidian with $k=2$ or $k=3$, and two elements are related via distance vector ($r_i = r_j + d_{ij}$). 
Often, such sequences are described in a continuous space as a parametrized function known as a curve. 
All previous sequence learning tasks are applicable to spatial sequences as well.
However, because elements have a distance relation the solutions to prediction, recognition, and generation tasks might leverage spatial structure.
For example, two vertical lines of equal length can have different locations, thus different sequences of numbers. 
Still, they have the same pairwise differences and thus are more similar than two random lines. 
Note, movement trajectories like in the thought experiment are spatial sequences.

Sequence learning arises in many domains and applications.
In neuroscience, as a sequential activation of spiking neurons, or chunking the perception into the sequence of events.
Recent work suggests that cells that encode places, boundaries, and landmarks arise from a pure sequence learning \cite{Raju2022}.
In machine learning, the statistical approach is used to learn sequences on large datasets.
Particularly, the LSTM and transformer models with many different variations are applied to text, sounds, or even images. 
Reinforcement learning deals with sequential decision-making and can be reframed as a large sequence learning task \cite{janner2021reinforcement}.
Sequence prediction is at the core of algorithmic information theory where the prediction heavily depends on the sequence complexity \cite{Hutter2007}.
In control theory, a fusion of visual and inertial information to track the robot's position in space can also be treated as sequence-to-sequence learning \cite{clark2017vinet}.
Note, that all these fields deal with slightly different flavors of sequence learning, whether it is the difference in elements (text, sounds, spatial locations), discrete or continuous case, large or small database, statistical or one-shot learning. 
Therefore, all have different mathematical descriptions and algorithms.

\section{Spatial sequence representations and their problems}
\subsection{Linear movement}
Returning to the thought experiment, the three figures consist of three linear segments.
To encode a line segment it is enough to specify the position of two endpoints $(p_0, p_1)$. 
The most common representation of a segment is parametric: $p(t) = p_0 + (p_1 - p_0)t, t \in [0, 1]$ where a vector $p(t)$ encodes all segment points, see fig. \ref{fig:spatial_sequences}a. 
To represent the segment in hardware it should be discretized, either by selecting a number of points $N$ or a discretization step of parameter $t$: $\Delta t = L / N$, where $L$ is the segment length. 
In any case, for two dimensions the result is a sequence of numbers, 
like  $s_1=[(1,2), (2,4)...(100, 200)]$, that can be visualized in a 2D figure, like in figure \ref{fig:spatial_sequences}a.

The problem with this representation is that Euclidean transformation (translation, rotation, and scaling) leads to a completely different sequence of numbers. 
But we would like to know that the two segments are actually the same, but moved.
In computer vision, this problem of alignment between two sets of points is known as point set registration and classically is solved with iterative algorithms (like the iterative closest point). 
However, humans use another strategy, they extract invariant features from two sequences and find correspondences in feature space. 
Many deep learning architectures also learn features and then search for correspondences, however, they require a lot of training data and poorly generalize.
See a recent review for different approaches to point set registration \cite{huang2021comprehensive}. 

Also, humans do not use end-to-end learning but rather perform unsupervised feature extraction,  where features can be in the form of subprograms \cite{lake2015human}. 
For example, we could write the segment of a line in the differential form $\frac{dp(t)}{dt} = q, p(0) = p_0, t \in [0, L]$, where $q$ is a unit vector in the direction from $p_0$ to $p_1$. Or in discretized version $p_{i+1} = p_i + q \dot \Delta t, i=0:L/ \Delta t$, that can be interpreted as an instruction: “start at point $p_0$ and repeat a small movement in the direction $q$ until the length L is reached”. 
Another conceptually different instruction is “to repeat the movement until the position is close enough to the endpoint $p_1$”. 
The first case requires measuring the length. 
The second needs to perceive the current location globally and compare it with the stored coordinate.  
In any case, the two translated lines have different locations, but the instruction for change remains the same.
As was suggested, humans may widely use instructions for geometry processing dubbed “language of geometry” \cite{amalric2017language}.

The weak spot of the instruction form is that given two line segments it is unknown if they intersect. Both instructions should be executed, intermediate results saved, and then checked for collisions. 
This is in contrast to visual imagination, where all points of two segments are constantly present. 
Or another weakness is the length of a segment. 
What units to use? Meters? 
Interestingly, people do not imagine a linear segment with a certain length, it has direction but not size.
The size appears as a ratio to other objects, therefore next we consider several linear segments. 

\subsection{Piecewise linear movement}
\begin{figure}
	\centering
     \includegraphics[width=\textwidth]{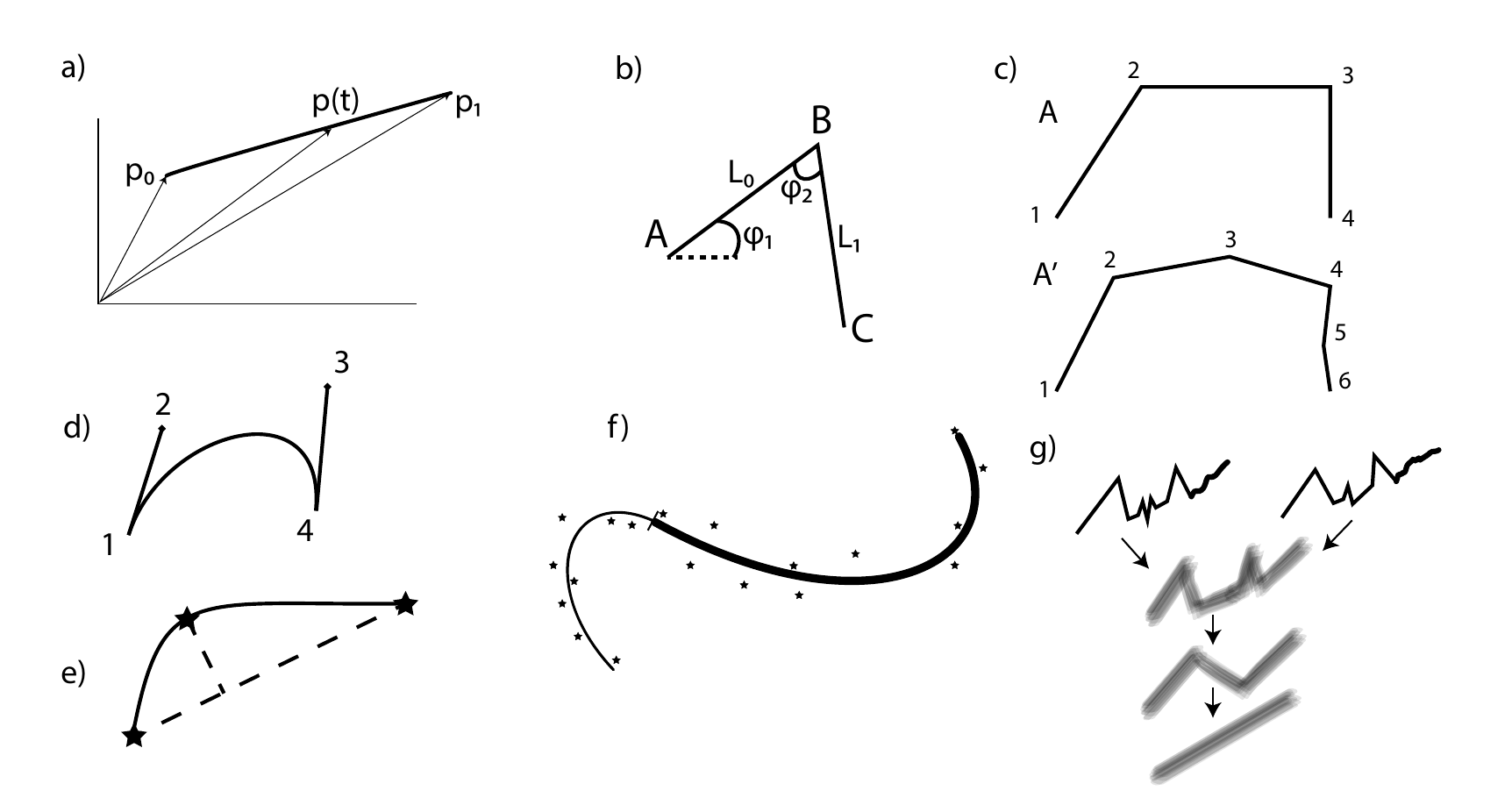}\hfill
	\caption{
 a)Parametric representation of a linear segment. 
 b)Representation of multiple segments with length and turning angles.
 c) Illustration that two similar spatial sequences may have very different representations.
 d) Bezier curve with handles.
 e) A curve as a shape with endpoints plus extremum point representation.
 f) Fitting points with two Bezier curves (shown with different thicknesses).
 g) Two different sequences are the same at a higher abstraction level.  
 }
	\label{fig:spatial_sequences}
\end{figure}

Two connected line segments can be represented as the coordinates of three key points (A, B, C) see fig. \ref{fig:spatial_sequences}b.
But again, spatial transformation changes the sequence of numbers completely.
A better way to represent segments as an initial position plus a sequence of turning angles and lengths, $p_0$ plus $(\phi_0, L_0), (\phi_1, L_1)$ (see a fundamental theorem of space curves). Here $\phi_0$ is an angle with a horizontal axis, and $\phi_1$ is relative to the first segment. 
Furthermore, to fix the problem of measuring length, we can specify not length but the ratio of segments length: $(\phi_1, a_1)$, where $a_1 = L_1/L_0$ and $L_0$ becomes some arbitrary length. 
Under rotation, turning angles do not change (except the first one), and the ratio stays the same under scaling. 

This representation is similar to what people use in navigation, they construct landmarks and learn instructions for going from one place to the next.  
However, consider fig. \ref{fig:spatial_sequences}c, two sequences are very similar in space but have different numbers of keypoint and slight variations that make comparison with turning angles representation complicated.
 So this chain-like relationship is not a robust strategy and 
 representation should include relations with other distant elements on different scales. 

\subsection{Curves}
In differential geometry, a curve is a continuous map from some interval $(a,b)$ to n-dimensional space $p: (a, b) \in R -> R^n$. The well known example is a 2d circle $p(t)= (r\cos t, r\sin t), t \in [0, 2\pi)$.
Similar to a linear segment, in computers, the compact functional form must be converted to a sequence of numbers. 

In practical applications like vector computer graphics, complex curves usually are approximated with the composition of Bezier curves. They are easy to manipulate, resolution-free, and have a compact functional representation compared to pixels in raster graphics. 
2D Bezier curve compresses a sequence of point positions to just 8 numbers that describe four control points, see fig. \ref{fig:spatial_sequences}d.
Many more representations of curves exist (see a book chapter \cite{gleicher2021curves}) but all trade computation over memory, that is store a long sequence of numbers as a function with a much shorter sequence of parameters.

But again, under transformation, noise, and variations these parameters change, and no invariances in the parameter space are preserved. 
Consider in contrast instruction representation of turning angles, similar to piecewise linear segment. 
The circle can be represented as: “1. start at some position; 2. move in some direction by a small amount; 3. turn by a small angle; 4. Repeat 2 and 3 until return to a starting position”.
This instruction holds for all circles at any position and any scale, with the only difference in the amount of movement.

 Experiments in human perception suggest that sensory inputs of the movement trajectories are also encoded in a compact form similar to programs \cite{AlRoumi2021}.
 This observation is linked with the minimum description length (MDL) principle, that neural networks encode the sequence with the shortest instruction, thus becoming more energy-efficient in terms of neural activation.   However, as we show next, shorter is not necessarily better.
 
\subsection{Real spatial sequences}
The key distinction of real spatial sequences, like movements on the skin,  striding pedestrians, or swinging of a falling leaf is that they are perceived in a form of raw data as the sequence of numbers (or receptor activation).  Instead of two points for a line, or parameters of the curve, we have just a sequence of numbers. 

The statistical approach proposes to use a parametric model and to optimize some criteria to fit parameters to the data. 
With this method, we can approximate the sequence of numbers with a composite Bezier curve, splines, or some other parametric function, like an artificial neural network. 
See for example fig.\ref{fig:spatial_sequences}f, where points are fitted with two Bezier curves, which makes a compact representation. 
However, similar to images, where objects will always produce variations in pixel values, real spatial sequences never repeat exactly the same trajectory. Thus, curve fitting is challenged with the same problem of generalization, how one model is applicable to different sequences. A huge amount of research is dedicated to this topic, where bias-variance tradeoff takes the central place. A good example of finding this tradeoff for curve fitting with B-splines is presented in the work \cite{Figueiredo2000}.

Recently, Vapnik, one of the founders of the statistical learning theory, introduced the notion of statistical invariants (as derived from predicates, see \cite{vapnik2020complete}), special equalities that link observations and the hypothesis of the model, like the parameters for the Bezier curve. 
These invariants are different from features from classic learning, they reduce the space of possible hypotheses, while more features increase the hypothesis space. 
With good invariants, learning is faster, requires less data, and gives better generalization.
But finding these invariants Vapnik calls the “intelligent part of the learning problem”, in contrast to finding parameters to fit the data. 

Such invariants may be a symmetrical relationship, or some other rules that link the features, like the relative position of eyes and a nose for a face. 
In previous sections, it was shown that the instructions  for curves that follow the movement of attention can provide the invariance for translation, rotation, and scaling. Perhaps, the movement of attention can help with finding good rules that explain and predict the sequence. 
However, in addition, four issues need to be addressed in the next section. 

\section{Movement of attention}
\subsection{Movement in space}
Consider again the thought experiment. The sequence of positions is perceived not element by element like music, but as a lasting perception of the overall shape. Most likely, frontoparietal connections are responsible for this, where the frontal cortex act as a working memory (with complex nested tree-like structure) for elements of the sequence represented in the parietal cortex, see more at \cite{wang2019representation}.  

Sequence learning applied to temporal sequences usually faces the problem of long-range dependencies, because of exponentially increased dimensionality of possible combinations (state space).  The temporal-to-spatial recoding for spatial sequences changes the problem of long-range dependencies to the search for relationships in the spatial domain. The movement of attention between persistent salient features gives the distance relationships and can provide additional invariances like symmetry or specific spatial patterns like the corners of a triangle. With the  movements, we can approximate a curve as a collection of three features:  the position and direction of two endpoints plus the position of the most distant point from the line that connects the endpoints (see fig.\ref{fig:spatial_sequences}e).  This represents a 1d curve more like a 2d shape and makes the task of spatial sequence learning similar to active vision, where the image is processed not as a whole but as a sequence of features from the eye movements. 

In addition, the spatial representation explicitly represents line-line intersection, which is not possible with the instruction form. This follows everyday experience, draw two lines on the skin, and you know if they intersect. Furthermore, the intersection point becomes salient and acts as a new feature to describe the composite shape. 
The same is true for visual imagination, mental movements of two line segments create at some moment an intersection, and we can estimate the relative distance from the intersection to the endpoint or any other spatial relationships. 

It is far from clear how exactly this space representation is formed on a neural level.
Perhaps, something functionally similar to grid cells, place cells, distance cells, etc. that are found in the hippocampus and entorhinal network is present in the neocortex (in our case most likely parietal), as was hypothesized in \cite{lewis2019locations}. Or perhaps the entorhinal cortex represents not just locations in space, but visual space as well, following some evidence provided in  \cite{killian2012map, julian2018human}. Nevertheless, it feels that mental travel in different locations is different from a mental rotation of a triangle.
So it is likely that a dedicated network should exist, but the neural mechanism for encoding space and visual-object space might be similar. 

\subsection{Movement is scale}
 If the analogy with the encoding space in the hippocampus is correct then we can make parallels with the scale as well. The dorsal part of the hippocampus encodes place on a small scale, like rooms, and the ventral part on a larger scale, like streets \cite{Harland17}. This dorsoventral longitudinal axis forms a scale axis. 
 We may  hypothesize that similarly a spatial sequence is encoded in different scales with varying accuracy of details.
 Two features, like corners, might be treated separately on one scale, but on the other as one feature.
 This directly relates to grouping by spatial proximity from Gestalt principles. 
 Consider two different curves (fig.\ref{fig:spatial_sequences}g) that perceptually look similar.
This similarity arises because the sequence representations are the same on the level of some neural populations responsible for encoding data at some scale. 
This hypothesized axis of scale for curves would encode data in a coarse-to-fine manner. 

Unfortunately, this mechanism is not found, and no computational algorithms can demonstrate such a bottom-up, unsupervised coarse-to-fine representation of spatial sequences yet. In 1984, this idea was demonstrated for 1d data, with the solution in a form of a tree where the deeper level encodes more fine features \cite{witkin1984scale}. Subsequent works on 2d data on images were not as successful but resulted in other algorithms known as active contours or snakes \cite{kass1988snakes}. 

Representation of spatial sequences on different scales results in different levels of abstraction. However, this forms a redundant representation, since the lowest level contains all information. This deviates from the conclusions made in recent paper \cite{AlRoumi2021}, that humans represent sequences following the MDL principle. It is possible that their conclusion is valid for studied spatial sequences of locations on the octagon. But we argue, that in order to compare different sequences and be able to generalize, redundant representation on multiple levels of abstractions is necessary. 

Similar confusion is present in the compression community, where the primary goal is to achieve lossless compression that later can always be made coarse, see \href{http://prize.hutter1.net/hfaq.htm#lossless}{M.Hutter web page}.  But finding similarity and patterns  in  a compressed representation is not trivial \cite{Mishra2016}.

Representation on multiple scales may be an instance of a balance in memory-computation trade-off where more memory (needed for redundant representation) is traded for less computation (to compare sequences and find similarities for generalization).  Classical point set registration algorithms do not use redundant memory for intermediate features and instead use more computations. 
Humans, most likely, have multiple representations that exist simultaneously, and attention selects the necessary level of description for a given task.
So attention moves not just in space but in scale, 
 group close features together, or differentiate more complex to composite. 

\begin{figure}
	\centering
	\includegraphics[width=\textwidth]{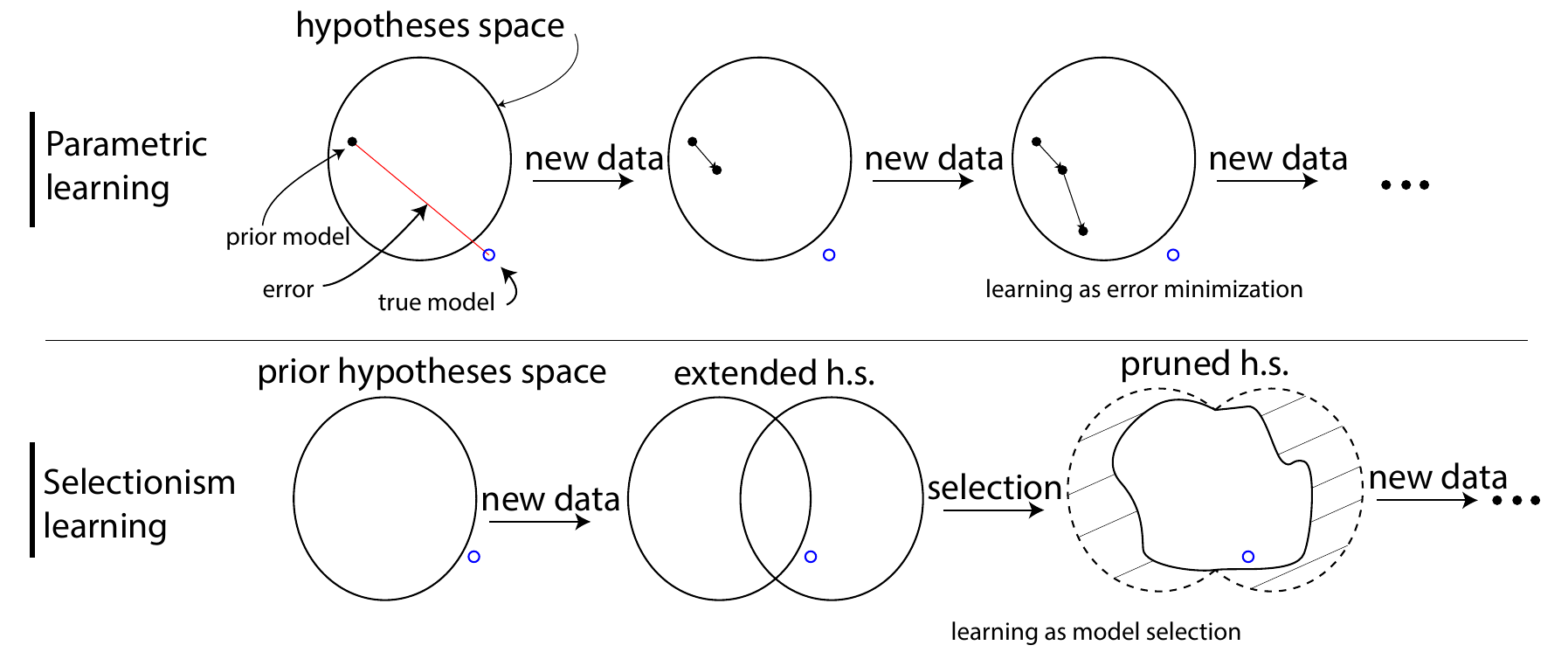}\hfill
	\caption{Illustration of the difference between parametric learning and selectionism learning. Top: the hypotheses space is fixed and learning is a  search for the best model. Bottom: }
	\label{fig:selectionism}
\end{figure}

\subsection{Movement in hierarchy}

Temporal-to-spatial recoding makes it easier to form spatial relationships with temporarily distant features. However, if the spatial sequence becomes too long, with too many features, the capacity of working memory in the frontal cortex is not enough to sustain spatial representation. The well-known strategy to fix this is to chunk, or combine, features into another composite feature that requires less working memory \cite{ramkumar2016chunking}. 

This changes the mathematical representation of the sequence to a recursive form 
$U = \{(s_i,r_i): s_i \in U, r_i \in R^k, i=1:N\}$ as a collection of features $s_i$ 
at some locations where each $s_i$ can be as well composed of feature-location pairs. 
Thus, a numerical sequence of positions is replaced with a sequence of symbols at some positions.
Moreover, each recursive level may be represented in a local reference frame,
very similar to the architecture widely adopted in computer graphics.   

An interesting option is that the parietal cortex stores these symbols as small chunks of typical patterns of the movement of attention.  Given that the parietal cortex has a similar structure to other parts, like visual or auditory, and we know that later store composite sensory features, it is  reasonable to assume that the parietal cortex does the same with the inputs of the movements of attention. And possibly its role not directly in attentional control, as suggested before \cite{shomstein2012cognitive}, but in attentional pattern recognition and completion. Moreover, similarly to the movement in scale, attention moves across a hierarchy of composite features to find the appropriate level for some task. How it is different from movement in scale is unclear, maybe the same mechanism is used with the exception that on a larger scale, details are not lost but compressed in the form of symbols. 

\subsection{Origin of the movements}
Sequence learning is based on the ability to compare sequences. In a simple case prediction or recognition can be done with a nearest-neighbor search. But how to form a good feature space to make comparisons, especially without large datasets and classical supervised or self-supervised setup? 

We propose two hypotheses that might lead to an answer.  

\textit{Hypothesis 1 (Selection)}. 
Good features, invariants, instructions, and levels of abstraction
can be received by a selection process from a large dynamic set of possible candidates.

This hypothesis is illustrated in fig.3, where learning is not a movement in a fixed parametric space to minimize some error function, but a generation and selection of the best hypotheses that explain and predict the data.
An example of such a generation process is multiple saccades of the eye that scan the scene to receive distance and direction relationships.
We hypothesize that similarly the movements of attention are made in an excessive manner but in a cognitive space, which also includes movement in scale, that groups or differentiates features. 
Relationships from movements that reoccur or correlate across different inputs are reinforced and less likely to be erased.
The selected relationships might form good compact representations for 
solving sequence learning tasks, particularly to perform comparison and sustain generalization. Such extensive search and elimination might be the mechanism for finding statistical invariants introduced by Vapnik.

This hypothesis is based on two previous ideas.
The first is selectionism, where the large population diversity increases the probability to have individuals with some useful qualities that will not be filtered by an external selection process.
It has many realizations in other fields, like evolution or optimization, and was promoted to explain brain functioning by neural group selection, known as "Neural Darwinism" \cite{edelman1987neural}.  
The proposed mechanisms of polychronization \cite{izhikevich2006polychronization} in spiking neural networks turned out to be wrong and cannot stably implement such selection.
Still, a modern view on cell assemblies formation with multiple inhibitory and disinhibitory circuits \cite{sadeh2021excitatory, osaulenko2021formation} might form a basis with sufficient memory capacity for such redundancy and selectivity. 

The second idea is from the models of computer vision before the deep learning surge (pre-2012). 
Many wrote about what deep learning brought new, but few mention what was unfortunately forgotten.
The formation of saliency maps from the first principles of information theory and the selection of features with further comparison were good brain-inspired ideas. 
However, they lost the race in the benchmark metrics, and instead of further research were overshadowed by the deep learning approach.  

A straightforward implementation of the proposed selection hypothesis using the pre-2012 approach fails. It quickly runs into computational limitations since the number of possible relationships grows exponentially. Therefore to select better relationships given a finite amount of memory we need to order locations and scales where to attend by importance. Also,  other heuristics will be useful, like the composition of features based on some criteria (see  Gestalt principles). However, selectionism does not propose to return to handcrafted features but to use lessons from a previous learning decade and utilize huge computational resources for generating and selecting good features. But to overcome exponential explosion we make a second hypothesis.

\textit{Hypothesis 2 (Nonparametric data structure)}.  An efficient data structure that compactly stores features and their relationships and allows for quick search, composition, and comparison exists and is not based on neural networks.

This hypothesis is based on the analogy to algorithms and data structures for string processing (like radix tree, suffix tree, etc).  They provide an inspiring example, where the search among a billion strings requires examining only around 30 bits. String symbols should not be fundamentally different from features or even instructions and reduction in computation should apply here as well. 

Such reduction is widespread in biological networks. On a local level, some inhibitory neurons target specifically axon hillock (spike generation site) to prevent neurons from sending action potential.  What is not sent usually may not be computed.  On a system level,  the activation function of every neuron in a network is not calculated, but only of certain areas selected by the movements of attention. 

 These two hypotheses invite deviating from the common approach of finding parameters that fit the data according to some loss function. And to explore nonparametric learning procedures based on selectionism with redundant movements of attention. 

 \section{Conclusions}
In this paper, with the help of a thought experiment,
we highlighted some properties of human spatial sequence perception. 
We argued that the invariance to transformation arises from the movement of attention, which is similar to the instruction form for spatial sequence representation.
Unlike temporal data, spatial data leave a persistent activation due to the working memory.
This temporal-to-spatial recoding alleviates the problem of long-range dependencies and helps to uncover good features, like symmetry.  
The movement of attention in scale is like the movement in a coarse-to-fine axis that reduces the complexity of a sequence but increases redundancy. 
This is in contrast to the MDL principle, but we argue that redundancy is inevitable in order to have multiple levels of abstraction that are the basis for generalization.  

We provided two hypotheses on how to make and organize the right movements of attention. Instead of finding iteratively the better solution in some parametric model, selectionism learning forms as many possible relationships and features  with more data, with further elimination of the less useful. To deal with computational complexity the second hypothesis purposes that some efficient data structures could store different features with a quick search and low memory. 
Still, many unresolved questions remain, like how to precisely define the usefulness of features without the error function, or what is a good problem setup to test the proposed hypotheses.

Overall, the topics mentioned in this paper like composition, hierarchy, generalization, and multiple levels of abstraction are well known and eagerly discussed in the literature, (see recent review  \cite{Whittington2022}), but no agreement is made on how to approach these problems. This paper proposes to explore the nonparametric selectionism approach to learning.

\bibliographystyle{abbrv}
\bibliography{references}  






\end{document}